\documentclass[sigconf]{acmart}

\renewcommand\footnotetextcopyrightpermission[1]{} 
\AtBeginDocument{%
  \providecommand\BibTeX{{%
    \normalfont B\kern-0.5em{\scshape i\kern-0.25em b}\kern-0.8em\TeX}}}

\usepackage{graphicx}
\usepackage{epstopdf}
\usepackage{footnote}
\usepackage{balance}
\usepackage{amssymb}
\usepackage{amsmath}
\usepackage{multicol}
\usepackage{multirow}
\usepackage{anyfontsize}
\usepackage{array}
\usepackage{hhline}
\usepackage{float}
\usepackage{tabularx}

\begin{document}

\title{Adapting Computer Vision Algorithms  \linebreak for Omnidirectional Video}

%
\author{Hannes Fassold}
\affiliation{%
  \institution{JOANNEUM RESEARCH, DIGITAL - Institute for Information and Communication Technologies}
  \streetaddress{Steyrergasse 17}
  \city{Graz}
  \country{Austria}
  \postcode{8010}
}
\email{hannes.fassold@joanneum.at}

\renewcommand{\shortauthors}{Hannes Fassold}

\begin{abstract}
  Omnidirectional ($360^\circ$) video has got quite popular because it provides a highly immersive viewing experience. For computer vision algorithms, it poses several challenges, like the special (equirectangular) projection commonly employed and the huge image size. In this work, we give a high-level overview of these challenges and outline strategies how to adapt computer vision algorithm for the specifics of omnidirectional video.

\end{abstract}

%
%


\maketitle


\section{Introduction}

\begin{figure}[t]
	\centering
		\includegraphics[width=0.43\textwidth]{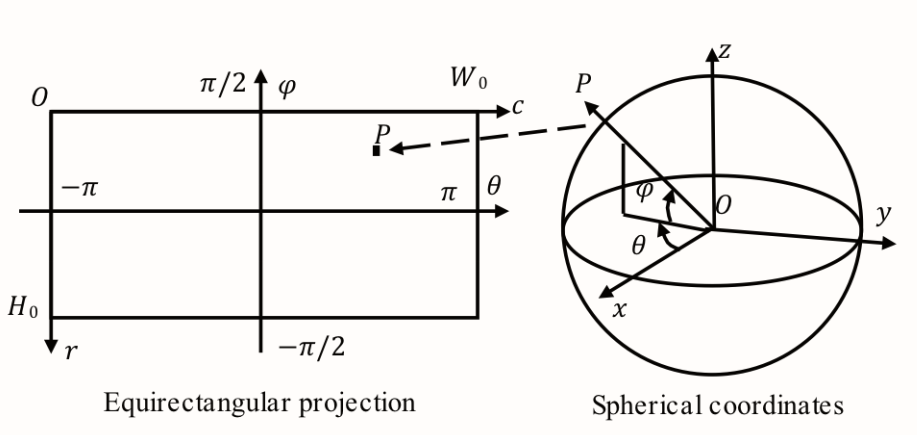}
	\caption{Coordinate systems employed for an omnidirectional image. Image courtesy of \cite{Suzuki2018}.}
	\label{fig:coordinate}
\end{figure}

Omnidirectional ($360^\circ$) video content recently got very popular in the media industry as well as in robotics, because it allows the viewer to experience the content in an immersive and interactive way. Omnidirectional consumer video cameras like the Samsung Gear 360 or the Ricoh Theta V have multiple lenses and capture images which cover the whole viewing sphere, typically in 4K or UltraHD resolution. Omnidirectional videos are typically consumed with a head-mounted display (HMD), so that the user is free to choose the area (viewport) within the sphere he is currently interested in. The whole viewing sphere is encoded in one 2D image for each timepoint, usually in equirectangular projection \cite{Lee2010}. Coordinates on the viewing sphere are usually given in a longitude-latidue representation (see Figure \ref{fig:coordinate} for the relation between the viewing sphere and the 2D image). In the following, the longitude is always denoted by $\phi$ and has the range $\left[-180, 180\right]$ degrees. The latitude is always denoted by $\theta$ and has the range $\left[-90, 90\right]$ degrees. 

Content captured with an omnidirectional camera poses several challenges for computer vision algorithms. Firstly, the captured images have a high resolution (usually UltraHD with 3,840 x 2,120 pixel) as they have to cover the whole viewing sphere. This leads to a high processing time of the algorithm unless some adaption strategies (like  spatial subsampling) are employed . Secondly, due to the equirectangular projection which is commonly employed, the areas of the sphere which are away from the equator are stretched in the image and the left and right image border represent adjacent regions in the viewing sphere. When applying an algorithm naively to the input image in equirectangular projection, it likely will produce non-optimal results.

Therefore, strategies are needed in order to adapt computer vision algorithms designed for content captured with a conventional camera to the specifics of omnidirectional video. These strategies can be grouped into two categories, which we will name \emph{viewport-centric processing} and \emph{image-centric processing}.  In the following, we will describe the two categories and outline the strategies which can be employed for each category.

\addtolength{\textheight}{-1cm}

\section{Viewport-centric processing}

In viewport-centric processing, the whole viewing sphere is split up into a set overlapping viewports $\{\tau_k\}$ so that the full sphere is covered. For each viewport $\tau_k$, the respective viewport image $v_k$ in rectilinear projection is rendered from the omnidirectional image $I$. The computer vision algorithm is now applied to each viewport image $v_k$, giving the result $r_k$ (e.g. a list of detected objects for an object detector or the denoised image for an image restoration algorithm). The set of per-viewport results $r_k$ have to be fused now into a single result $R$ and back-projected into an image in equirectangular projection. 

\begin{figure}[t]
	\centering
		\includegraphics[width=0.43\textwidth]{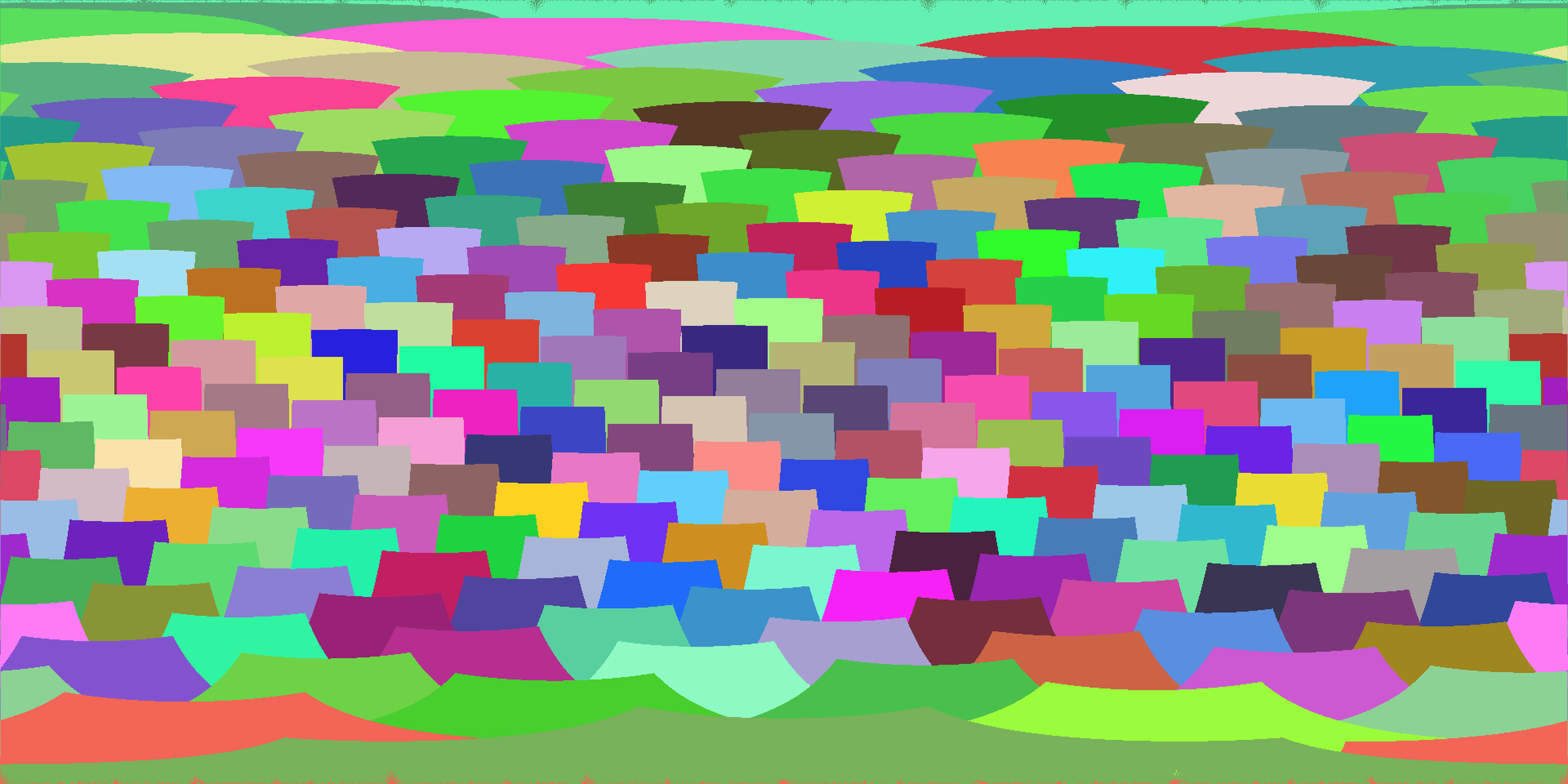}
	\caption{Visualization of 240 overlapping viewports employed in viewport-centric processing (back-projected to equirectangular projection).}
	\label{fig:viewports}
\end{figure}

For an object detector (like YoloV3 \cite{Redmon2018}), the fusion step usually involves some sort of non-maximum suppression, in order to suppress multiple detections occurring at the same region of the viewing sphere. For an image restoration algorithm where the result is the enhanced image, the fusion and back-projection step can be done via some sort of image blending. Specifically, the blending can be done with an accumulator image and weight image in a similar fashion as described in \cite{Rosner2010} for the image warping algorithm. For the generation of the overlapping viewports, the \emph{Vogel method} \cite{Swinbank1999} can be employed for generating approximately uniformly distributed points on the viewing sphere, which will serve as the center of the respective viewport. In Figure~\ref{fig:viewports} a decomposition of the viewing sphere into $240$ overlapping quadratical viewports with a FOV of $24^\circ$ is visualized.

For viewport-centric processing, the computer vision algorithm can be applied \emph{without} major adaptions as the viewport image $v_k$ has been rendered in the usual rectilinear projection employed in the pinhole camera model. On the other hand, the decomposition into overlapping viewports means that significantly more pixels have to be processed compared with image-centric processing. Of course, the viewport images have also to be rendered (and back-projected) in a reasonable resolution which takes a certain amount of time, even on the GPU. Furthermore, a good strategy for the fusion step (which is task-specific) has to be researched and implemented.

\section{Image-centric processing}


In image-centric processing, the input image from the omnidirectional video is processed by the specific computer vision algorithm as a whole, without the intermediate step of rendering a set a viewport images. This means that the existing computer vision algorithm has to be adapted to the peculiarities of omnidirectional images, specifically the equirectangular projection commonly employed and the adjacency (in the viewing sphere) of the left and right image border regions. Without these adaptions, the quality of the algorithm may be significantly worse or even completely unusable.

Processing the input image as a whole is usually much more efficient than viewport-centric processing, as runtime-intensive steps like the viewport rendering and the final back-projection are not necessary and less pixels have to be processed. Furthermore, many computer vision algorithm nowadays are GPU-accelerated, and GPU-acceleration is usually more effective (higher speedup factor compared to CPU) for larger images. 

The kind of adaption strategy which is employed depends heavily on the components employed within the computer vision algorithm. E.g. for the image blur measure proposed in \cite{Marziliano2002}, the global blur magnitude is calculated from a statistical analysis of the widths of vertical image edges. In order to adapt this algorithm properly for an input image in equirectangular projection, one has to compensate for the stretching of vertical image edges near the top and bottom border of the image (the poles of the viewing sphere). This can be done by calculating a distortion map with the local stretching factors and multiply the measured edge width with a compensation factor derived from the distortion map. 

\begin{figure}[t]
	\centering
		\includegraphics[width=0.43\textwidth]{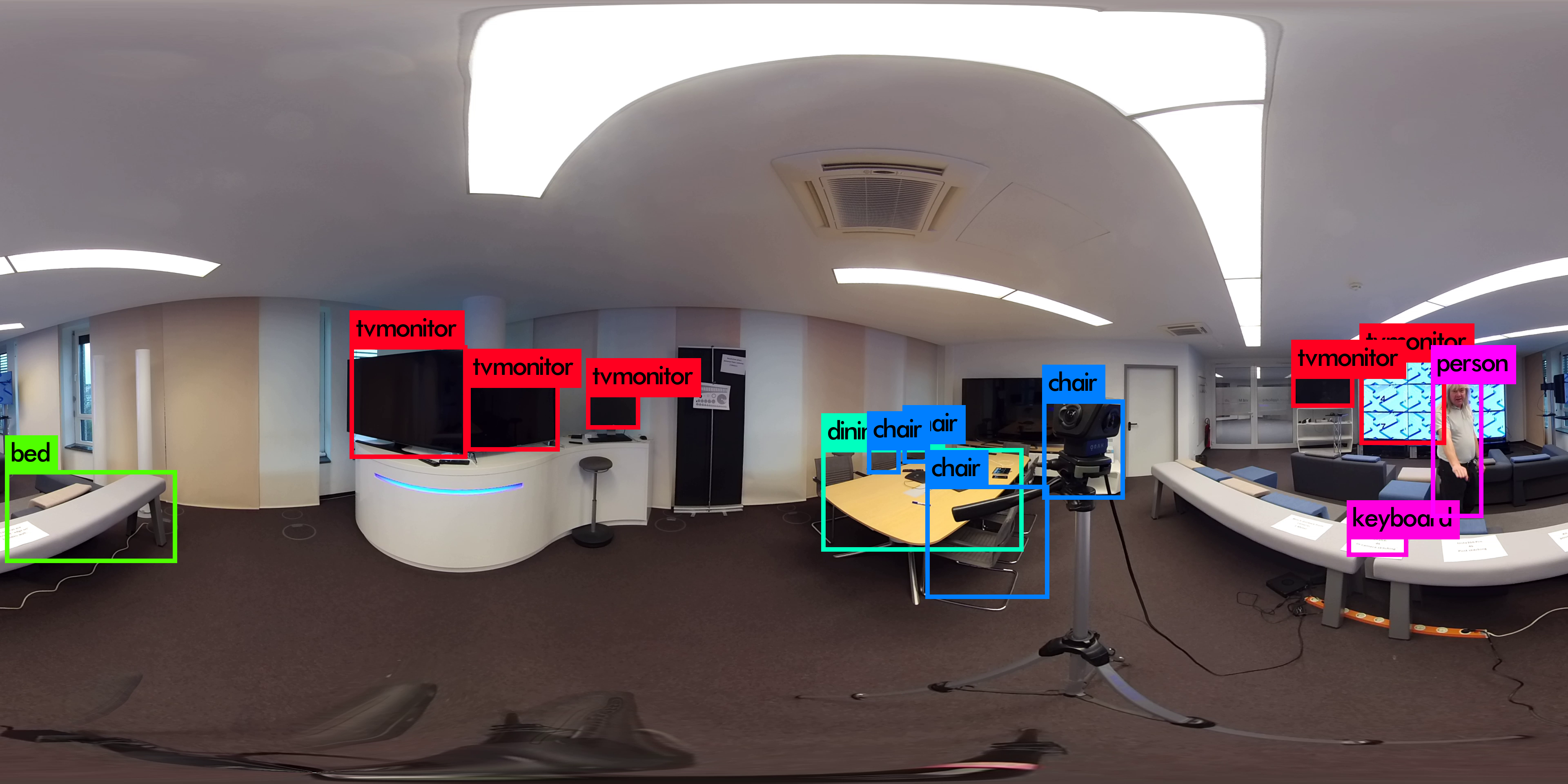}
	\caption{Result of object detector YoloV3 \cite{Redmon2018} for an omnidirectional image.} 
	\label{fig:yolov3detection}
\end{figure}

State of the art object detectors like YoloV3 \cite{Redmon2018} seem to be quite robust against the stretching induced by the equirectangular projection, but are prone to multiple detections of the same object if it appears partially at the left and right image border. A possible adaption strategy for that case would be to merge these double detection via some sort of non-maximum suppression. The non-maximum suppression must take into account of course that the longitude $\phi$ is cyclic. An exemplary result of the YoloV3 object detector for an image in equirectangular projection can be seen in Figure \ref{fig:yolov3detection}. The receptive field of the object detector was set to $896\times448$ pixel in order to account for the $2 : 1$ aspect ratio of the input image.

\section{Conclusion}

In this work, we described the challenges omnidirectional video poses for computer vision algorithms designed for conventional video. Approaches for adapting these algorithms like viewport-centric processing or image-centric processing were outlined, and the advantages / disadvantages of each approach were discussed.






\begin{acks}
This work has received funding from the European Union's Horizon 2020 research and innovation programme, grant n$^\circ$ 761934, Hyper360 (``Enriching 360 media with 3D storytelling and personalisation elements'').

\end{acks}

\bibliographystyle{ACM-Reference-Format}
\bibliography{sample-base}

\end{document}